%% file: iclr2026_conference.tex
\documentclass{article} 
\usepackage{iclr2026_conference,times}
\usepackage{booktabs}
\usepackage{tabularx}
\usepackage{makecell}
\usepackage{array}
\usepackage{colortbl}
\usepackage{xcolor}
\usepackage{graphicx} 
\usepackage{booktabs}
\usepackage{tabularx}
\usepackage{subcaption}   
\usepackage{array}
\input{math_commands.tex}

\usepackage{xspace}
\usepackage{hyperref}
\usepackage{url}
\usepackage{wrapfig}
\usepackage{pifont}
\usepackage{subcaption} 

\newcommand{\minisection}[1]{\noindent{\textbf{#1}.}}

\newlength\savewidth

\definecolor{citecolor}{RGB}{34,139,34}
\definecolor{lightred}{RGB}{241,140,142}
\definecolor{amber(sae/ece)}{rgb}{1.0, 0.49, 0.0}
\definecolor{battleshipgrey}{rgb}{0.52, 0.52, 0.51}
\definecolor{cadmiumorange}{rgb}{0.93, 0.53, 0.18}
\definecolor{applegreen}{rgb}{0.55, 0.71, 0.0}
\definecolor{cadmiumgreen}{rgb}{0.0, 0.42, 0.24}
\definecolor{forestgreen}{rgb}{0.13, 0.55, 0.13}
\definecolor{red}{rgb}{0.89, 0.0, 0.13}

\definecolor{uscgold}{HTML}{d9ae02}      
\definecolor{carnegiered}{HTML}{C41230} 
\definecolor{berkeleyblue}{HTML}{003057}

\newcommand{\NAME}{\textsc{DAVE}\xspace}

\title{\NAME: A VLM Vision Encoder for Document Understanding and Web Agents}

\author{%
  \textbf{Brandon Huang\textsuperscript{1,2}},\;
  \textbf{Hang Hua\textsuperscript{1}},\;
  \textbf{Zhuoran Yu\textsuperscript{1,3}},\;
  \textbf{Trevor Darrell\textsuperscript{2}},\;\\
  \textbf{Rogerio Feris\textsuperscript{1,\dag}},\;
  \textbf{and Roei Herzig\textsuperscript{1,\dag}}\\[4pt]
  \textsuperscript{1}MIT-IBM Watson AI Lab,
  \textsuperscript{2}UC Berkeley,
  \textsuperscript{3}University of Wisconsin–Madison
}

\begin{document}

\maketitle
\let\thefootnote\relax\footnotetext{\textsuperscript{\dag}Equal advising. Correspondence to: \texttt{zhaobin@berkeley.edu, hang.hua1@ibm.com}}
\input{sec/abstract}
\input{sec/intro}
\input{sec/related_works}
\input{sec/method}
\input{sec/experiment}
\input{sec/results}
\input{sec/conclusion}
\input{sec/limitations.tex}



 \bibliography{iclr2026_conference}
 \bibliographystyle{iclr2026_conference}

\newpage
\input{sec/appendix}

\end{document}

%% file: math_commands.tex
\usepackage{amsmath,amsfonts,bm}

\def\eqref#1{equation~\ref{#1}}

\def\1{\bm{1}}

\DeclareMathAlphabet{\mathsfit}{\encodingdefault}{\sfdefault}{m}{sl}
\SetMathAlphabet{\mathsfit}{bold}{\encodingdefault}{\sfdefault}{bx}{n}



%% file: sec/abstract.tex
\begin{abstract}
While Vision–language models (VLMs) have demonstrated remarkable performance across multi-modal tasks, their choice of vision encoders presents a fundamental weakness: their low-level features lack the robust structural and spatial information essential for document understanding and web agents. To bridge this gap, we introduce \NAME, a vision encoder purpose‑built for VLMs and tailored for these tasks.  
Our training pipeline is designed to leverage abundant unlabeled data to bypass the need for costly large‑scale annotations for document and web images.
We begin with a self-supervised pretraining stage on unlabeled images, followed by a supervised autoregressive pretraining stage, where the model learns tasks like parsing and localization from limited, high-quality data.
Within the supervised stage, we adopt two strategies to improve our encoder's alignment with both general visual knowledge and diverse document and web agentic tasks: 
(i) We introduce a novel model-merging scheme, combining encoders trained with different text decoders to ensure broad compatibility with different web agentic architectures.
(ii) We use ensemble training to fuse features from pretrained generalist encoders (e.g., SigLIP2) with our own document and web-specific representations. 
Extensive experiments on classic document tasks, VQAs, web localization, and agent-based benchmarks validate the effectiveness of our approach, establishing \NAME as a strong vision encoder for document and web applications.
\end{abstract}

%% file: sec/intro.tex
\section{Introduction}

Vision-Language Models (VLMs)~\citep{liu2024llavanext,bai2025qwen2,achiam2023gpt} have shown remarkable capabilities in multi-modal reasoning and understanding, enabling a wide range of applications from image captioning to interactive web agents~\citep{wu2024atlas, qin2025ui, liu2023llava, li2022blip}. A central component of VLMs is the vision encoder~\citep{yin2024survey}, which converts images into visual tokens for the language backbone to process~\citep{li2023blip, liu2023llava}. As such, the design of these encoders has become a key focus in VLM research~\citep{tong2024cambrian, shi2024eagle}.

However, the vision encoders predominantly used by VLMs suffer from a fundamental weakness: their low-level features lack the robust structural and spatial information essential for document understanding and web agents~\citep{tschannen2025siglip, radford2021learning}. Conversely, DINO‑style models~\citep{oquab2023dinov2} offer low-level features yet are tuned to natural images~\citep{oquab2023dinov2,simeoni2025dinov3} and transfer poorly to documents, UIs, and diagrams~\citep{tong2024cambrian}. 
To bridge this gap, we introduce \NAME (\emph{\textbf{D}ocument and web \textbf{A}gents \textbf{V}ision \textbf{E}ncoder}), a vision encoder purpose‑built for VLMs and tailored for these tasks. See Figure~\ref{fig:fig_main} for an overview.

A key challenge in training on document and web images is the scarcity of high-quality annotated data, since current annotations rely on OCR models that bottleneck both scalability and quality~\citep{kim2021donut, lee2023pix2struct}. We address this with a two-stage training process: self-supervised pretraining on large-scale unlabeled data, followed by autoregressive pretraining on limited annotated structural and grounding data. Yet, the supervised stage introduces a further challenge: specialized vision encoders trained with a single text decoder tend to overfit to that decoder, leading to misalignment when paired with other decoders. We mitigate this issue via multi-decoder pretraining and weight-merging (“model soup”)~\citep{wortsman2022model}, which produces a largely decoder-agnostic encoder. To preserve high‑level semantics while leveraging our structural and spatial features, we fuse a generalist encoder with our encoder during pretraining via ensembling. The resulting specialized encoder integrates seamlessly with diverse decoders and generalist VLM stacks. 
Finally, our purpose-built vision encoder serves as the core component of a VLM architecture specifically designed for document and web agent tasks.
We evaluate \NAME across diverse settings, including traditional and vision–language tasks. On traditional benchmarks, \NAME surpasses all SOTA models in document recognition and segmentation, while achieving competitive results with SigLIP~2 on screenshot classification. For vision–language evaluation, we test on document understanding, web localization, and web agent. Notably, \textbf{\NAME improves performance by an average of 10.5\% compared to SigLIP~2}, highlighting its ability to provide specialized representations. Finally, we evaluate web agent settings on the Mind2Web benchmark~\citep{deng2023mind2web}, where \NAME improves web agent performance by 5\% over the strongest baseline vision encoder.

\input{fig/fig_main}

We summarize our main contributions as follows: \textbf{(i)} We introduce \NAME, a vision encoder purpose‑built for VLMs and tailored for document and web agents tasks;
\textbf{(ii)} We propose a two-stage pretraining framework that combines a self-supervised pretraining stage on unlabeled images, followed by a supervised autoregressive pretraining stage;
\textbf{(iii)} We introduce model weight merging and ensemble training strategies to enhance the alignment of structural and spatial representation for documents and web images with diverse VLM and agentic frameworks;
\textbf{(iv)} We perform extensive experiments to evaluate the impact of visual representations on document and web understanding, as well as downstream web-agent tasks.  

%% file: fig/fig_main.tex
\begin{figure}[t]
    \centering
    \includegraphics[width=1\linewidth]{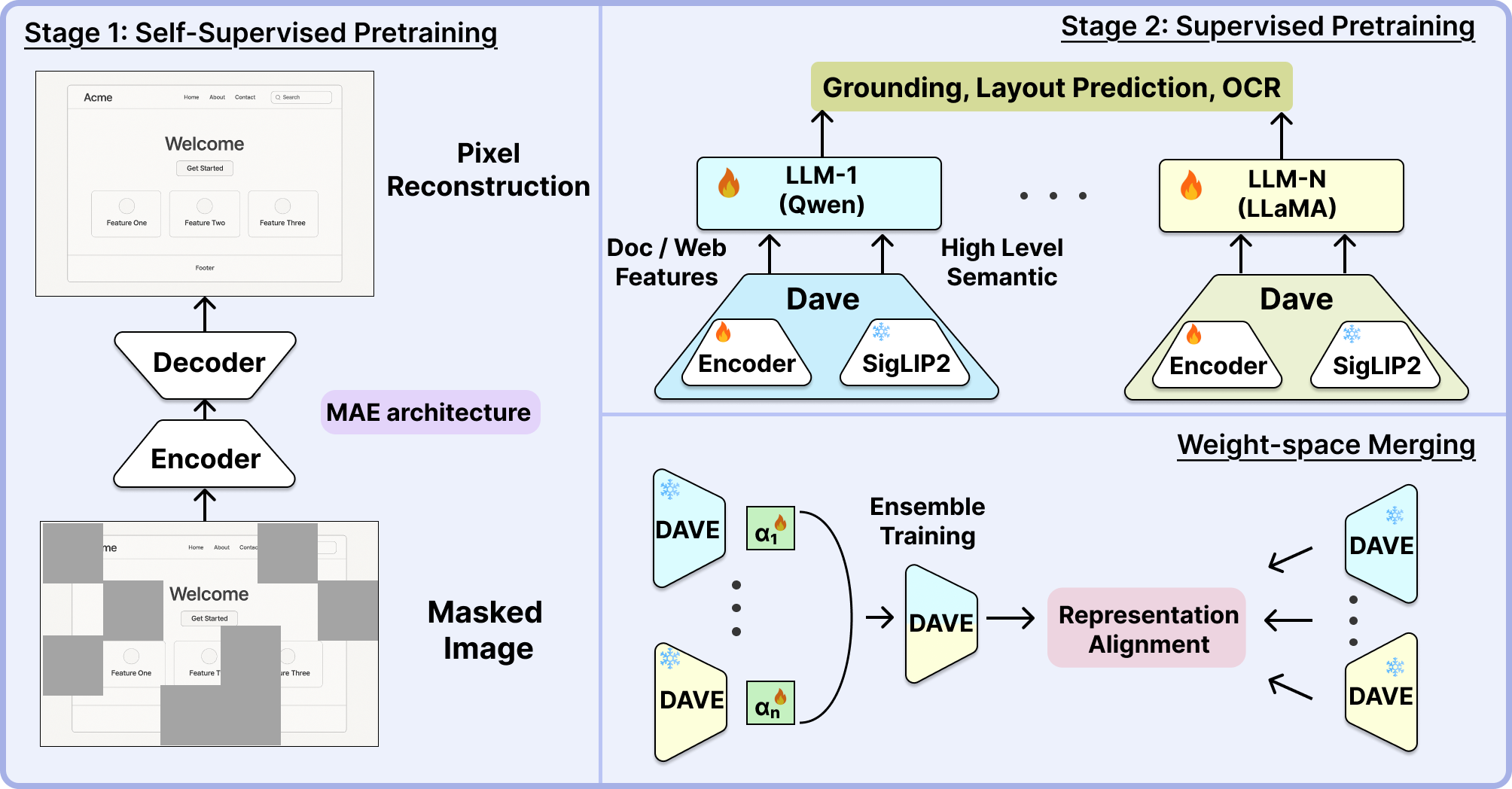}
    \caption{
\textbf{\NAME Overview}.
Stage 1 trains the vision encoder with a decoder using MAE, learning strong structural and spatial priors from unlabeled data. Stage 2 performs autoregressive pretraining on diverse tasks with different text decoders and fuses the high-level semantic features from SigLIP 2. After that, different encoders are combined into a single one by learning a merge coefficient using unsupervised representation alignment, while keeping the encoders frozen.
    }
    \label{fig:fig_main}
\end{figure}

%% file: sec/related_works.tex
\section{Related Work}

\minisection{Pretrained Vision Encoders} 
Image–text contrastive learning~\citep{radford2021learning, zhai2023sigmoid, jia2021scaling, yu2022coca} and self-supervised representation learning~\citep{he2022masked, zhou2021ibot, oquab2023dinov2, chen2020simple, caron2021emerging} are the two main paradigms for large-scale visual pretraining. Recent work has sought to combine the strengths of the two approaches through joint training or post-hoc alignment~\citep{tschannen2025siglip, fini2025multimodal, maninis2024tips, naeem2024silc}, yielding strong encoders for both vision and multimodal tasks. The works most closely related to ours are Eagle ~\citep{shi2024eagle} and Perception Encoder ~\citep{bolya2025perception}, which align pretrained vision encoders to a pretrained LLM with large-scale general datasets. In contrast, we pretrain the vision encoder from scratch with self-supervised learning, use multiple pretrained LLMs as a backbone, and use domain-specific data to create a specialized foundation model.

\minisection{Document Understanding with LLMs}
Document processing and understanding have transitioned from OCR-based~\citep{xu2020layoutlm, xu2020layoutlmv2} methods to vision-language-based. One line of work trains encoder-decoder models with various objectives, like extracting structures~\citep{feng2025dolphin, kim2022ocr, lv2023kosmos, lee2023pix2struct}, and image reconstruction conditioned on text~\citep{huang2022layoutlmv3, tang2023unifying}. Recent advancements in VLMs led to another line of work that uses document data to finetune a VLM~\citep{nassar2025smoldocling, hu2024mplug, liu2024textmonkey} or incorporate it as part of the VLM training~\citep{team2025granite, bai2025qwen2, chen2024expanding}. Our work seeks to bridge these two lines of work by introducing a specialized vision encoder within a general VLM framework.

\minisection{Vision-Language Models}
Inspired by the success of recent large language models (LLMs)~\citep{brown2020language,dubey2024llama}, Vision-language models~\citep{radford2021learning,liu2024llavanext,sun2025latent,hua2024v2xum,ye2023mplug,tang2024avicuna,bai2025qwen2,zhu2025internvl3,hua2025finecaption,tong2024cambrian} aim to achieve multimodal intelligence by jointly understanding and generating visual and language information. 
Flamingo~\citep{alayrac2022flamingo} and BLIP-2~\citep{li2023blip} are two of the early works that explore integrating LLMs as part of VLM. Beginning with LLaVA~\citep{liu2024llavanext}, researchers have used instruction-following chat data in VQA format for instruction tuning, achieving significantly improved results~\citep{liu2024mmbench,li2023seed,hua2024mmcomposition, tang2025vidcomposition,lu2023mathvista}.

\minisection{Multimodal Web Agents}
Vision-based autonomous web agents have recently attracted attention for their simplicity and stronger generalization compared to LLM-based agents~\citep{deng2023mind2web, kim2023language}. Early efforts such as WebGUM~\citep{furuta2023multimodal} and CogAgent~\citep{hong2024cogagent} pretrained vision–language models (VLMs) on web and GUI data to enhance agentic capabilities. Follow-up work~\citep{wu2024atlas, xu2024aguvis, liu2024harnessing, qin2025ui} expanded training with large-scale grounding and interaction datasets, further improving VLM performance on web-based tasks. While these works are promising steps toward a general-purpose visual web agent, the role of the vision encoder in these systems remains underexplored.

%% file: sec/method.tex
\section{The DAVE Model}

\subsection{Preliminarilies} 
\label{sec:prelim}

\minisection{Masked Autoencoder}
Masked Autoencoder (MAE)~\citep{he2022masked} adopts an asymmetric encoder–decoder architecture. The vision encoder processes only a subset of visible patches, while a lightweight transformer decoder reconstructs the full image from the encoded features and mask tokens. After pretraining, only the encoder is retained for downstream tasks.  

By default, MAE applies per-patch normalization before computing the reconstruction loss to learn better representations. The reconstruction loss is then defined as
{\small
\begin{equation}
\mathcal{L}_{\text{MAE}} = \frac{1}{|\mathcal{M}|} \sum_{i \in \mathcal{M}}
\left\| f_\theta(\tilde{x})_i - \frac{x_i - \mu(x_i)}{\sqrt{\sigma^2(x_i) + \epsilon}} \right\|_2^2,
\quad \epsilon = 10^{-6}
\label{eq:mae}
\end{equation}
}
\noindent where $\tilde{x}$ denotes the input image with masked patches, $f_\theta(\tilde{x})_i$ is the reconstructed output of patch $i$ predicted by the decoder with parameters $\theta$, $x_i$ is the ground-truth pixel values of patch $i$, $\mu(x_i)$ and $\sigma^2(x_i)$ are the mean and variance of pixels within patch $i$, $\mathcal{M}$ is the set of masked patches.

\minisection{Vision--Language Model}  
A vision--language model (VLM) consists of a vision encoder $\phi$, an MLP projector, and a text decoder $\Theta$. The vision encoder takes an input image $x \in \mathbb{R}^{H \times W \times 3}$ and produces a sequence of patch-level features $\{v_i\}_{i=1}^N$. These are projected into the text embedding space via the MLP projector and used along with the text input.  

In what follows, we describe the process to get our vision encoder. 

\subsection{Stage~1: SSL on Document and Web Images}
\label{sec:stage_1}
In Stage~1, we conduct self-supervised training with MAE to learn rich structural and spatial features from document and web images. While MAE demonstrates the strongest performance on OCR tasks compared to other methods, the training is prone to instability when trained at scale~\citep{fan2025scaling}. We find that this instability is particularly acute for document and web images. Our analysis (Section~\ref{sec:analysis}) attributes this to their characteristically low inter-patch variance, which destabilizes the target normalization in the standard MAE objective (Equation~\ref{eq:mae}). To address this, we modify the objective to reconstruct raw pixel values directly:
\begin{equation}
\mathcal{L}_{\text{MAE-pixel}} = \frac{1}{|\mathcal{M}|} \sum_{i \in \mathcal{M}}
\left\| f_\theta(\tilde{x})_i - x_i \right\|_2^2,
\label{eq:mae-pixel}
\end{equation}
This change stabilizes training and enables scaling the training sample to 20 million images without additional hyperparameter tuning. With this robust self-supervised procedure, we then proceed to the second stage of our training process.

\subsection{Stage~2: Supervised Multi-Task Pretraining}
To further enhance the encoder's structural and spatial understanding for document and web images, we perform a supervised, autoregressive pretraining stage. This stage utilizes a limited set of high-quality labeled data for tasks like OCR, layout extraction, and web localizations. 
We use the VLM architecture as described in Section~\ref{sec:prelim}, and \NAME as the vision in the VLM architecture.

\minisection{Weight-space Merging}
\label{sec:stage_1}
A key limitation of the architecture discussed above is that the pretrained vision encoder becomes tightly coupled to its specific text decoder. This coupling significantly degrades performance when the encoder is integrated with a different decoder. To address this, we use model merging to create a vision encoder that is agnostic to the choice of text decoder.

Formally, given a set of $n$ pretrained text decoders $\{\Theta_1, \dots, \Theta_n\}$, we train $n$ corresponding instances of our vision encoder, denoted $\{\phi_1, \dots, \phi_n\}$. Each instance is aligned with a different text decoder but otherwise shares the same architecture, training data, and hyperparameters.

To merge these different encoders, we propose a distillation-based merging scheme that learns a small set of coefficients to combine pretrained weights while keeping the original parameters frozen.  
Considering each encoder $\phi_i$ as a set of $m$ weights $\{\theta_i^{(j)}\}_{j=1}^m$, we form each merged weight $\theta_{\text{merge}}^{(j)}$ by learning a set of coefficients $\{\alpha_i^{(j)}\}_{i=1}^n$ to compute the weighted sum of the corresponding weight for each of the $n$ encoders:
\[
\theta_{\text{merge}}^{(j)} = \sum_{i=1}^n \alpha_i^{(j)} \theta_i^{(j)}
\qquad \alpha_i^{(j)} \in [0, 1].
\]
The merged encoder is thus composed of $m$ merged weights: $ \phi_{\text{merge}} = \{\theta_{\text{merge}}^{(j)}\}_{j=1}^m$.

The resulting encoder produces patch-level features $\mathbf{z} = \phi_{\text{merge}}(\mathcal{I})$. To ensure that $\phi_{\text{merge}}$ preserves the features from all $n$ teacher encoders, we define a distillation loss, $\mathcal{L}_{\text{distill}}$. This objective minimizes the average Mean Squared Error between the merged features $\mathbf{z}$ and the features from each teacher encoder $\mathbf{z}_i = \phi_i(\mathcal{I})$:
\[
\mathcal{L}_{\text{distill}}
= \frac{1}{n} \sum_{i=1}^n
\left\| \hat{\mathbf{z}}_i - \mathbf{z}_i \right\|_2^2.
\]
During the distillation process, all encoder parameters remain frozen; only the newly introduced combination coefficients are optimized. The final vision encoder is given by the merged encoder:
\[
\phi_{\text{\NAME}}^{\text{final}} = \phi_{\text{merge}}(\{\alpha_i^{\star}\}_{i=1}^n),
\]
where $\{\alpha_i^{\star}\}_{i=1}^n$ are the optimized coefficients obtained from this training.

The next section details our method for fusing specialized document and web features with general visual representations.

\minisection{Ensemble Training}
Pretraining the encoder exclusively on document and web data provides strong structural and spatial features, but it also limits its grasp of general visual representation. This is a crucial shortcoming, as the high-level semantic features learned from diverse, large-scale datasets are equally important for robust performance.

To obtain both types of features, we design an ensemble pretraining paradigm that combines a frozen pretrained generalist encoder $\phi_{\text{gen}}$ with our document and web specialist encoder $\phi_{\text{spec}}$ from the previous stage: $ \phi_{\text{DAVE}}(x) = \text{Concat }\!\big(\phi_{\text{gen}}(x), \, \phi_{\text{spec}}(x)\big).$


This design provides two main benefits: (i) it encourages $\phi_{\text{spec}}(x)$ to focus on learning low-level structural and spatial representations, as high-level semantics are captured by $\phi_{\text{gen}}(x)$; and (ii) it enables early fusion of structural and spatial features with high-level semantic features.

%% file: sec/experiment.tex
\section{Evaluation}
\subsection{Implementation Details}
\minisection{Self-supervised Stage}
For the SSL stage, we follow the original MAE implementation \citep{he2022masked}, with ViT-L16-384 \citep{dosovitskiy2020image} as our vision encoder. We train the MAE with a batch size of 4096 for 120K steps. More detail can be found in Appendix~\ref{appx:mae}

\minisection{Supervised Training Stage}
We adopt the VLM architecture discussed in Section~\ref{sec:prelim}, with an image tilting size of 384. During the ensemble training, we employ frozen SigLIP-2 as the generalist vision encoder, while the encoder from the Self-supervised stage serves as the domain-specialized component. This forms the full \NAME encoder.
We experiment with multiple LLMs of varying scales and architectures, including QWen2.5-0.5B-Instruct \citep{bai2025qwen2}, Phi-4-mini-Instruct \citep{abouelenin2025phi}, and Granite-3.1-3B-Instruct \citep{team2025granite}. After a single epoch of full-parameter training, we retain only the \NAME encoder for weight-merging and downstream tasks. During the weight-merging distillation, we train the merge coefficient for 20 epochs on unlabeled documents and web images. Refer to Appendix~\ref{ssl:impl} for implementation details.

\subsection{Evaluation Setting}
In this section, we first describe our encoder baselines, followed by two types of evaluations for the vision encoders: (i) finetuning vision encoders for classic document tasks, (ii) performing instruction tuning to build a VLM with the vision encoder for vision-language tasks and agentic tasks.

\minisection{Baselines}
\label{exp:baseline}
We compare our vision encoder with both generalist and specialist encoders. For generalist encoders, we utilize DinoV2 \citep{oquab2023dinov2} and Web-SSL MAE \citep{fan2025scaling}, a MAE variant trained on 2 billion images. We also include SigLIP2 \citep{tschannen2025siglip} and AIMv2 \citep{fini2025multimodal}, which are SOTA encoders trained with both contrastive and reconstruction. For specialist vision encoders, we compare against DiT~\citep{li2022dit}, Pix2Struct \citep{lee2023pix2struct}, and Dolphin \citep{feng2025dolphin}. Since both Pix2struct and Dolphin are encoder-decoder models, we use their encoders for comparison.

\minisection{Classic Document Tasks}
For each classic document task, we finetune the vision encoder and the suitable prediction heads. In DocBank, we use attention pooling to pool the visual feature, followed by an MLP head to predict the bounding box. In Doclaynet, we train an MLP head to predict the semantic segmentation based on the feature patches. For RICO-SCA, we use attention pooling followed by an MLP to predict the class. Refer to Appendix~\ref{appx:cls_doc} for more details.

\minisection{Vision-Language Model as Evaluator}
To evaluate our vision encoders on more complicated and realistic tasks, such as VQA and instruction-based localization, we use VLMs as evaluators. Specifically, we follow the standard LLaVA architecture as discussed in Section~\ref{sec:prelim}. We use a tilting size of 336 for AIMv2 and 384 for all other vision encoders. For vision encoders that produce a different number of visual tokens than the tilting resolution, we use bilinear interpolation as in LLaVA-Onevision~\citep{li2024llava} to interpolate the visual tokens. Note that we omit the projector alignment phase to improve training efficiency, as prior work has shown that this stage yields only limited gains~\cite{tong2024cambrian, karamcheti2024prismatic}. We use Llama-3.2-3B-Instruct~\citep{dubey2024llama} and Qwen2.5-7B-Instruct~\citep{bai2025qwen2} as the LLM backbone. We train for one epoch with the vision encoder frozen. For Mind2web, we finetune the VLMs on the training set before performing offline evaluation on the test set. More detail can be found in Appendix~\ref{appx:vlm_eval}.

\input{table/tab1}
\subsection{Benchmarks}
In this section, we describe our evaluation datasets for several downstream benchmarks.

\minisection{Classic Document Understanding Tasks}
To directly evaluate the structural and spatial visual representation from the vision encoders, we consider several document parsing tasks that can be performed with only images. We employ DocBank~\citep{li2020docbank} to comprehensively evaluate document recognition on different categories, including tables, charts, and paragraphs. We also use Doclaynet~\citep{pfitzmann2022doclaynet} to test document segmentation. For web tasks, we use RICO-SCA~\citep{li2020mapping} to perform web UI classification.

\minisection{Document and General VQA}
For document understanding and question answering, we evaluate on AI2D \citep{Kembhavi2016}, OCRBench \citep{Liu_2024}, DocVQA \citep{mathew2021docvqa}, InfoVQA \citep{Mathew_2022_WACV}, and ChartQA \citep{masry2022chartqa}. For general vision-language understanding, we report results on MMMU \citep{yue2024mmmu}, RealWorldQA \citep{xai2024realworldqa}, and TextVQA \citep{singh2019textvqa}. 

\minisection{Web UI and Agent Tasks}
For web UI localization, we use Screenspot-V2 \citep{wu2024atlas}, which consists of text instructions (e.g., ``click the button in cooridnate (x, y)''), and the corresponding bounding boxes. We report the center accuracy: the label-box center lies within the predicted box. We also evaluate on WebSRC~\citep{chen2021websrc} and VisualWebBench~\citep{liu2024visualwebbench} for web UI question answering. For the agentic benchmark, we evaluate on Mind2Web~\citep{deng2023mind2web}, where the model receives a user goal and a webpage screenshot, and must predict an action like clicking.

\subsection{Datasets}

In this section, we provide a description of our training datasets for the different stages.

\input{table/tab2}
\minisection{Self-supervised Learning Data}
We use 20 million images for our self-supervised training. For document data, we sample 10 million PDF images from DocFM~\citep{team2025granite}, which contains 85 million document pages extracted from Common Crawl, Wikipedia, and ESG (Environmental, Social, and Governance) reports. The PDFs are filtered to include English only. For web screenshot, we sample 10 million from Common-Web~\citep{commonscreens} without any language filtering.

\minisection{Supervised Learning Data}
For the autoregressive supervised training, we leverage data from diverse sources, including PlotQA \citep{methani2020plotqa}, ChartQA \citep{masry2022chartqa}, fintabnet~\citep{zheng2021global}, Datikz~\citep{belouadi2023automatikz}, Pubtables~\citep{smock2022pubtables}, and DocFM \citep{team2025granite}.  Importantly, we use the version from Granite Vision~\citep{team2025granite} where all the problems are reformulated into content extraction. The data for this stage is a mixture of curated benchmarks and a large-scale, self-processed dataset. The curated benchmarks cover tasks such as chart-to-markdown, table-to-caption, and web UI grounding with UGround~\cite{gou2024navigating}. To expand on this, we processed 500K PDF samples from arXiv with an OCR model~\citep{cui2025paddleocr}, formulating them into additional recognition and grounding tasks. Altogether, this training stage comprises approximately 2 million samples.
\input{table/tab3}

\minisection{Instruction Tuning Data}
Following Cambrian-1~\citep{tong2024cambrian}, We use the LLaVA-1.5-mix 665K~\citep{liu2023llava}, DocVQA \citep{mathew2021docvqa}, ChartQA \citep{masry2022chartqa}, and AI2D \citep{Kembhavi2016} as our instruction tuning data. We also integrate Pixmo-Doc \citep{yang2025scaling} into the data mix to improve document understanding. Additionally, we add MultiUI \citep{liu2024harnessingwebpageuistextrich} to improve the web and agentic capability. This combined to a total of 2.5 million training images.

%% file: table/tab1.tex
\newcolumntype{N}{>{\centering\arraybackslash}X}  
\newcolumntype{M}{>{\raggedright\arraybackslash}p{3.0cm}} 

\definecolor{mygreen}{rgb}{0.1, 0.6, 0.1}

\renewcommand{\arraystretch}{1.0}
\setlength{\tabcolsep}{2.5pt}

\begin{table*}[ht]
\centering
\footnotesize
\begin{tabularx}{\textwidth}{
  M
  *{11}{N}  
}
\toprule
& \multicolumn{5}{c}{\textbf{Document}}
& \multicolumn{3}{c}{\textbf{General VQA}}
& \multicolumn{3}{c}{\textbf{Web}} \\
\cmidrule(lr){2-6}
\cmidrule(lr){7-9}
\cmidrule(lr){10-12}
\textbf{Model / Variant}
& \textbf{\rotatebox{60}{AI2D}}
& \textbf{\rotatebox{60}{OCRBench}}
& \textbf{\rotatebox{60}{DocVQA}}
& \textbf{\rotatebox{60}{InfoVQA}}
& \textbf{\rotatebox{60}{ChartQA}}
& \textbf{\rotatebox{60}{MMMU}}
& \textbf{\rotatebox{60}{RealWorldQA}}
& \textbf{\rotatebox{60}{TextVQA}}
& \textbf{\rotatebox{60}{VisualWeb}}
& \textbf{\rotatebox{60}{Screenspot-V2}}
& \textbf{\rotatebox{60}{WebSRC}} \\
\midrule


\rowcolor[HTML]{e9edf6}
\multicolumn{12}{c}{\textbf{Llama-3.2-3B-Instruct}}\\
\midrule
MAE-Web                & 53.0& 27.6& 43.8& 26.2& 43.8& 29.9& 48.6& 35.1&36.9 & 51.1& 46.8\\
DinoV2           & 53.2& 2.6& 13.7& 20.4& 14.1& 35.0& 49.4& 13.7& 13.2& 16.5&17.6 \\
DiT               &49.9 & 2.1& 11.3& 19.2& 12.6& 28.8& 43.9& 10.0& 9.2& 2.5& 14.9\\
Pix2Struct       & 51.0& 6.9& 22.8& 21.4& 21.5& 33.3& 45.0& 16.4& 22.6& 32.6& 26.5\\
Dophine         & 50.8& 44.7& \underline{74.8}& 37.4& \underline{60.3}& 33.3& 44.4& 42.8& 48.8& \underline{56.3} & \underline{76.0} \\
AIMv2         & \underline{56.1}& 48.3& 56.6& 35.6& 48.7& 36.3& 51.4& \underline{58.7}& 46.7& 32.0& 49.4\\
SigLIP 2         & 58.0& \underline{51.5}& 72.1& \underline{40.6}& 51.8& \textbf{36.9} & \underline{53.5}& 64.4& \underline{54.7}& 40.7& 67.8 \\
\rowcolor{violet!10}
\textbf{\NAME (ours)}     &\textbf{59.6} & \textbf{62.2}& \textbf{82.1}& \textbf{50.2}& \textbf{63.1}& \underline{36.6}& \textbf{55.6}& \textbf{69.2}& \textbf{59.2}& \textbf{64.5}&\textbf{82.6} \\
\midrule

\rowcolor[HTML]{e9edf6}
\multicolumn{12}{c}{\textbf{Qwen-2.5-7B-Instruct}}\\
\midrule
MAE-Web        & 63.8& 33.2& 55.1& 28.6& 50.3& 38.4& 45.4& 40.6&59.5 & 70.1&60.7 \\
Pix2struct    & 58.7& 10.0& 29.8& 20.0& 28.3& 37.9& 39.1& 18.2& 52.3&47.8 &30.3 \\
Dophine      & 63.3& 50.9& 83.3& 42.0& 66.3& 40.2& 44.6& 49.4& 60.7& 75.2&85.6 \\

AIMv2         & 69.7& 69.3& 89.3& 57.8& 77.4& 41.3& 56.2&73.1 & 64.9& 79.3&86.1 \\
SigLIP 2     & \underline{73.6}& \underline{67.4}& \underline{90.2}& \underline{55.2}& \underline{75.9}& \underline{43.9}&  \textbf{58.7}&  \textbf{74.3}& \underline{65.4}&\underline{81.4} &\underline{88.3} \\
\rowcolor{violet!10}
\textbf{\NAME (ours)} &\textbf{74.0} & \textbf{67.5}& \textbf{90.9}& \textbf{60.2}& \textbf{82.5}&\textbf{45.8} &\underline{55.2} &\underline{73.7} & \textbf{67.3}&\textbf{82.9} &\textbf{88.6} \\
\bottomrule
\end{tabularx}
\caption{The \NAME's performance on Document understanding, general VQA, and Web understanding benchmarks with two VLM architectures using different LLMs. The best result per row is highlighted in \textbf{bold} and the second best with \underline{underline}. Higher values represent better performance.}
\label{tab:tab1}
\vspace{-4mm}
\end{table*}

%% file: table/tab2.tex
\newcolumntype{N}{>{\centering\arraybackslash}X}      
\newcolumntype{M}{>{\raggedright\arraybackslash}X}    

\renewcommand{\arraystretch}{1}
\setlength{\tabcolsep}{2pt} 

\begin{table*}[t]
\centering
\footnotesize
\begin{tabularx}{\textwidth}{@{} l *{6}{N} @{}}
\toprule
& \multicolumn{6}{c}{\textbf{Mind2Web}} \\
\cmidrule(lr){2-7}
\textbf{Model}
& \multicolumn{2}{c}{\textbf{Cross-Task}}
& \multicolumn{2}{c}{\textbf{Cross-Website}}
& \multicolumn{2}{c}{\textbf{Cross-Domain}} \\
\cmidrule(lr){2-3}\cmidrule(lr){4-5}\cmidrule(lr){6-7}
& \makecell{\textbf{Element}\\\textbf{Acc.}}
& \makecell{\textbf{Step}\\\textbf{SR}}
& \makecell{\textbf{Element}\\\textbf{Acc.}}
& \makecell{\textbf{Step}\\\textbf{SR}}
& \makecell{\textbf{Element}\\\textbf{Acc.}}
& \makecell{\textbf{Step}\\\textbf{SR}}
 \\
\midrule
\rowcolor[HTML]{e9edf6}
\multicolumn{7}{c}{\textbf{Llama-3.2-3B-Instruct}}\\
\midrule

MAE         & 21.0& 16.8& 16.5& 11.9& 15.5&11.6 \\
DinoV2        & 13.5& 11.4& 10.2& 6.6& 9.5&7.2 \\
DiT           & 5.8& 4.1& 2.7&1.2 &2.1 &1.1 \\
Pix2Struct           & 15.8& 12.3&12.1 &8.2 & 10.8&7.5 \\
Dolphin                 & \underline{24.0}& \underline{19.6}&\underline{20.9} &\underline{13.6} &\underline{19.3} &\underline{14.5} \\
AIMv2       & 14.7&11.6 & 9.2&5.9 &8.8 &6.1 \\
SigLIP 2         & 20.6& 17.3& 11.8& 8.7& 12.8&9.7 \\
\rowcolor{violet!10}
\textbf{\NAME (ours)}           & \textbf{ 30.8}& \textbf{ 26.1}& \textbf{ 24.2}& \textbf{ 18.0}&\textbf{ 23.9} &\textbf{ 19.1} \\
\bottomrule
\end{tabularx}
\caption{\textbf{Results on Web Agent.} Performance on Mind2Web with three splits (Cross-Task, Cross-Website, Cross-Domain). We report the stepwise accuracy (correct grounding) and the element accuracy (correct grounding and action) for each task.}
\label{tab:tab2}
\vspace{-3mm}
\end{table*}

%% file: table/tab3.tex
\begin{wraptable}{r}{0.5\textwidth}  
\centering
\footnotesize
\begin{tabular}{lccc}
\toprule
\textbf{Model}
& \textbf{DocLayNet}
& \textbf{DocBank}
& \textbf{RICO-SCA} \\
\midrule
DinoV2     & 68.4 & 38.3 & 85.6 \\
MAE-Web        & 64.6 & 44.5 & 88.3 \\
Dolphin    & 53.8 & 50.5 & 88.4 \\
Pix2Struct & 56.7   & 47.2   & 90.1   \\
AIMv2      & 70.5 & 42.4 & 91.3 \\
SipLIP 2   & 70.8 & 51.7 & \textbf{ 93.3} \\
\rowcolor{violet!10}
\textbf{\NAME}     &\textbf{ 74.1} & \textbf{ 56.9} & \underline{92.8} \\
\bottomrule
\end{tabular}
\caption{Performance comparison on \textbf{classic document tasks}.  
DocLayNet and DocBank use mAP, while RICO-SCA uses classification accuracy.}
\label{tab:tab3}

\end{wraptable}

%% file: sec/results.tex
\section{Results}
\subsection{Vision-Language Tasks}
Table~\ref{tab:tab1} shows the evaluation results for document, general, and web benchmarks. Overall, \NAME consistently outperforms the strongest baseline, SigLIP~2 in the Llama-3.2-3B-Instruct setup on 8 document and web benchmarks by an average of 10.5\%. This is achieved without losing the general VQA capabilities, like MMMU and RealWorldQA, which implies that the ensemble training successfully merges the structural and spatial features with the general visual features. In addition, when using Qwen-2.5-7B-Instruct as the VLM decoder, we show that the improvements in benchmarks hold a similar trend. This suggests that our merging scheme effectively aligns with different text decoders. Finally, one interesting observation is that specialized models like Pix2Struct and Dolphin perform suboptimally compared to SigLIP~2 and AIMv2, which shows that both general visual features and document/web-specific features are critical to the performance. This hypothesis is further supported by \NAME's strong results.
\input{fig/fig2.tex}

\subsection{Web Agentic Tasks}
The results for multimodal-mind2web are outlined in Table~\ref{tab:tab2}. \NAME outperforms the best baseline encoder, Dolphin, by an average of 5\%. Surprisingly, self-supervised vision encoders like MAE and DinoV2 perform competitively with SigLIP~2 and AIMv2, while document-specialized encoder Dolphine achieves the best accuracy among all baselines. This suggests that structural and spatial aspects may be more important in web navigation compared to broad general features.

\subsection{Classic Document Tasks}
As shown in Table~\ref{tab:tab3}, we evaluate \NAME on dense document tasks.
\NAME outperforms both specialized and generalist encoders on DocBank and Doylaynet. This demonstrates the strong structural and spatial visual features learned by our 2-stage pretraining. Furthermore, \NAME performs competitively with SigLIP~2 on screen classification, a task that emphasizes semantic understanding and for which \NAME is not designed to tackle.

\subsection{Ablations}
\label{abl}
 In this section, we present detailed ablations on the design choice of our training method. In all of the ablations, we report the average performance on document and web vision-language benchmarks, which we denote as Doc and Web in the Table~\ref{tab:tab5}.
 
\minisection{Training Method}
To assess the contribution of each design choice in the stage 2 supervised pretraining, we incrementally add each choice and plot the performance trajectory on document vision-language tasks tasks in Figure~\ref{fig:fig2}. Beginning with the a scratch text decoder and subsequently incorporating a pretrained LLM as the text decoder, ensemble training, and finally weight merging.
The consistently improving trend highlights the positive impact of our proposed design choices.

\minisection{Image Inter-Patch Variance}
\label{sec:analysis}
To show the instability of the MAE training on document and web image discussed in Section~\ref{sec:stage_1}, we plot the distribution density of inter-patch variance across different data sources in Figure~\ref{fig:fig3}. Compared to ImageNet~\citep{deng2009imagenet}, document and web images exhibit much lower variance. This distributional gap underscores the need for a specialized vision encoder capable of handling such data. Additional implementation details are provided in Appendix~\ref{appx:add_abl}.


\input{table/tab4}

\minisection{Different Merge Setting}
As described in Section~\ref{sec:stage_1}, we use weight merging to create a decoder-agnostic vision encoder.
Table~\ref{tab:tab5a} examines the effect of different merge strategies. Our end-to-end approach for learning merge coefficients outperforms both simple averaging and the heuristic-based Fisher Merge~\citep{matena2022merging} method, which weights parameters by their estimated importance using the Fisher information. Table~\ref{tab:tab5b} evaluates merging \NAME from different LLM backbones, where performance improves consistently with the number of merged encoders.

\minisection{Multi-encoder Comparison}
Because \NAME combines with both generalist and specialist features, we ask how \NAME performs compared to multi-encoder settings. Specifically, we concatenate different document and web specialized models with SigLIP~2 during the instruction tuning of building the VLM. In Table~\ref{tab:tab5c}, we present the result of different multi-encoder settings. We highlight that without any feature fusion, all the specialized encoders fail to improve performance on document and web benchmarks.

\minisection{Comparison with Finetuning}
To test whether our pretraining approach is more effective than directly finetuning an existing vision encoder on the supervised data used in our pretraining, we finetune SigLIP~2 with our supervised pretraining data. In Table~\ref{tab:tab5d}, we show that \NAME outperforms finetuned SigLIP~2, demonstrating the effectiveness of our approach.

%% file: fig/fig2.tex
\begin{figure}[t]
    \centering
    \begin{minipage}[t]{0.48\textwidth}
        \centering
        \includegraphics[width=\linewidth]{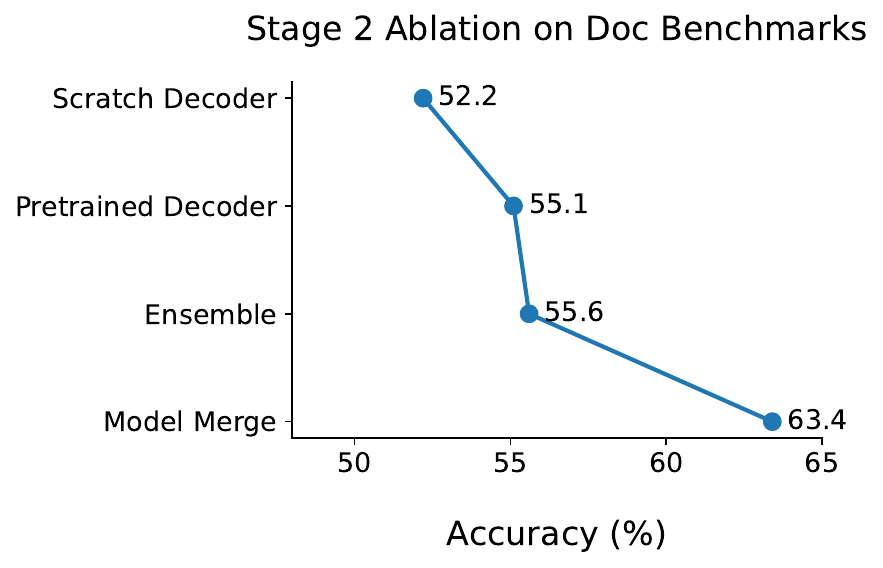}
        \caption{Each row indicates an additional modification to the training strategy.}
        \label{fig:fig2}
    \end{minipage}\hfill
    \begin{minipage}[t]{0.48\textwidth}
        \centering
        \includegraphics[width=\linewidth]{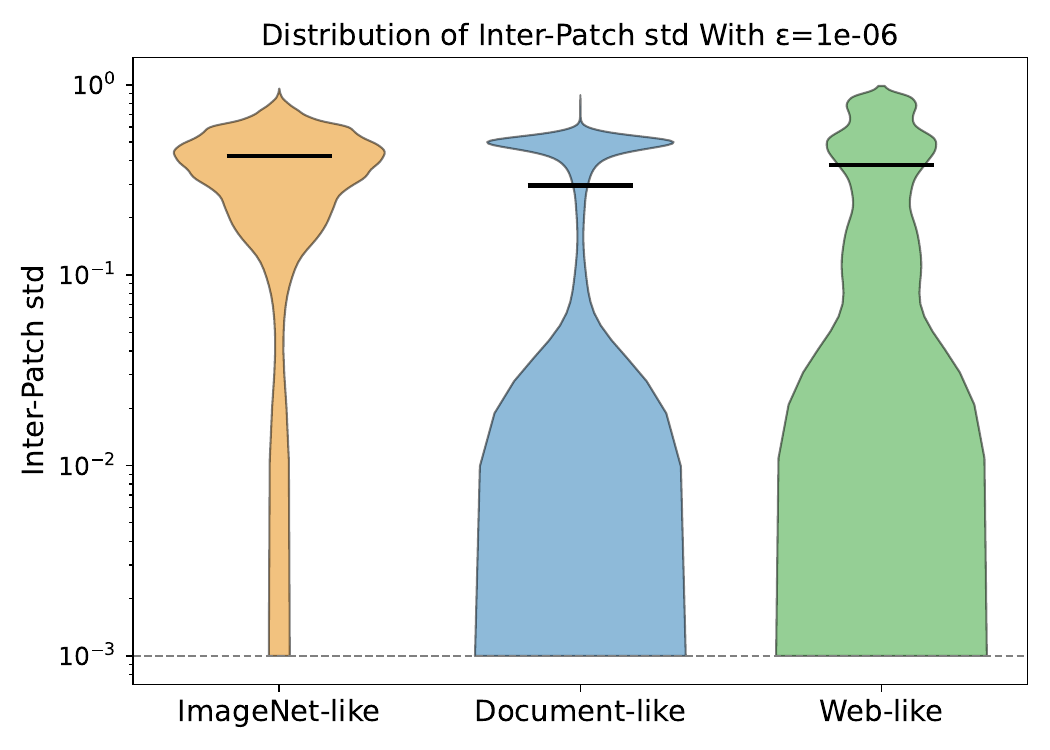}
        \caption{Inter-patch standard deviation across different data sources.}
        \label{fig:fig3}
    \end{minipage}
\end{figure}

%% file: table/tab4.tex
\newcolumntype{Y}{>{\centering\arraybackslash}X}
\newcolumntype{L}{>{\raggedright\arraybackslash}p{0.35\linewidth}} 
\newcolumntype{S}{>{\centering\arraybackslash}p{0.10\linewidth}}   

\begin{table*}[ht]
\centering
\footnotesize

\begin{minipage}{0.9\textwidth}
\centering
\begin{subtable}[t]{0.45\textwidth}
\centering
\begin{tabularx}{\linewidth}{L *{2}{Y}}
\toprule
 \textbf{Method}& \textbf{Doc} & \textbf{Web} \\
\midrule
No merge           & 55.6 &  53.0    \\
Average            & 62.8 &   67.7   \\
Fisher Merge       & 60.3 & 67.0     \\
Learned Coef & 63.4 &  68.2 \\
\bottomrule
\end{tabularx}
\caption{Comparison of Different Merge Methods.}
\label{tab:tab5a}
\end{subtable}\hfill
%
\begin{subtable}[t]{0.5\textwidth}
\centering
\begin{tabularx}{\linewidth}{L *{2}{Y}}
\toprule
\textbf{Method} & \textbf{Doc} & \textbf{Web} \\
\midrule
From Scratch        &   52.2   &   54.7   \\
Granite             & 55.6 &   53.0   \\
Granite+Qwen        &   62.1   & 64.8     \\
Granite+Qwen+Phi    & 63.4 &  68.2    \\
\bottomrule
\end{tabularx}
\caption{Comparison of Different Merging LLMs.}
\label{tab:tab5b}
\end{subtable}
\end{minipage}

\vspace{3em} 

\begin{minipage}{0.9\textwidth}
\centering
\begin{subtable}[t]{0.45\textwidth}
\centering
\begin{tabularx}{\linewidth}{L *{2}{Y}}
\toprule
\textbf{Method}& \textbf{Doc} & \textbf{Web} \\
\midrule
SIgLIP 2 + DiT  &   50.3   &    47.6  \\
SIgLIP 2 + PS  &   48.9   &   42.2   \\
SIgLIP 2 + DP  &   49.7   &   44.0   \\
\NAME  &   63.4   &   68.2   \\
\bottomrule
\end{tabularx}
\caption{Comparison of Different Specialized Encoder Fusion. Key: PS-Pix2struct, DP-Dolphin}
\label{tab:tab5c}
\end{subtable}\hfill
%
\begin{subtable}[t]{0.50\textwidth}
\centering
\begingroup
\setlength{\tabcolsep}{5pt}
\begin{tabularx}{\linewidth}{
  >{\raggedright\arraybackslash}X
  >{\centering\arraybackslash}p{0.2\linewidth}
  >{\centering\arraybackslash}p{0.2\linewidth}
}
\toprule
\textbf{Method} & \textbf{Doc} & \textbf{Web} \\
\midrule
SigLIP~2                 & 49.1 &   54.4   \\
Finetune SigLIP~2        &  58.2    &  65.2    \\
\NAME  &   63.4   &   68.2   \\

\bottomrule
\end{tabularx}
\endgroup
\caption{Comparison with Finetuning.}
\label{tab:tab5d}
\end{subtable}

\end{minipage}

\caption{\textbf{Ablation study}: (a) merge methods, (b) LLM merging, and additional analyses (c, d).}
\label{tab:tab5}
\end{table*}

%% file: sec/conclusion.tex
\section{conclusion}
We introduced \NAME, a foundational vision encoder for document and web understanding. Our approach combines self-supervised learning on large-scale unlabeled data with supervision from limited vision–language annotations, enabling data-efficient training in specialized domains. By leveraging pretrained generalist encoders, \NAME can focus on learning domain-specific features. Extensive experiments demonstrate that strong visual representations are critical for document, web, and agentic tasks, where our encoder achieves significant improvements over prior models. Ablation studies further confirm the method’s flexibility on different VLM and agentic frameworks.

%% file: sec/limitations.tex
\section{Limitations and Future Work} 
\label{sec:limitations}
Despite the significant progress achieved in document and web understanding by \NAME, several limitations remain. The model operates at a fixed resolution, depends on an external pretrained generalist encoder, and requires substantial compute for self-supervised training. In agentic settings, its vision encoder is also not conditioned on prior actions. These limitations highlight opportunities for developing vision encoders that are more compute- and data-efficient, can process inputs of any resolution or aspect ratio, and incorporate prior agentic actions to produce more tailored representations. In addition, this paradigm offers a promising path for domains with scarce vision–language data, such as medical imaging. While we do not anticipate specific negative impacts, as with any machine learning method, caution should be exercised during deployment.

%% file: sec/appendix.tex
\appendix
\section{Stage 1: Self-Supervised Training}
Here, we present the implementation detail and dataset used for the self-supervised training.
\subsection{Implementation details}
\label{appx:mae}
\input{appdx_items/ssl/ssl_hparams.tex}
We use the official implementation of masked autoencoder (MAE) for our training, and we use  32 H200 GPUs for the training. Our vision encoder is a ViT-L-384 initialized from scratch, whereas our decoder is a transformer model with a depth of 4 and 16 attention heads. The detail training hyperparameter can be found in Table ~\ref{tab:ssl-hparams}
\subsection{Datasets}
Our self-supervised pretraining dataset consists of 10 million document images from DocFM and 10 million web screenshots from Common Screen

\minisection{DocFM}
DocFM is a large scale document data collected by IBM consisting of 85 million document pages extracted from unique PDF documents sourced from Common Crawl, Wikipedia, and ESG (Environmental, Social, and Governance) reports. From this dataset, we first filter and keep only the English pdfs, then we randomly sample 10 million pdfs and store them as images.

\minisection{Common Screen}
Common Screen is a large scale web screenshot data consisting of 70 million screenshot images based on the Common Crawl data. We randomly sample 10 million images as our training data, without any language filtering.

\subsection{Additional Results on MAE}
\input{appdx_items/ssl/ssl_result.tex}

In this section, we demonstrates the effectiveness of pretraining on document and web images by comparing our stage-1 encoder, which we denote as MAE-Doc, with the the pretrained MAE-Web introduced in ~\ref{exp:baseline} in classic document tasks. In Table~\ref{tab:ssl-result}, the superior performance of MAE-Doc over MAE-Web highlights importance of the domain-specific pretraining for document and web tasks.

\subsection{Training Divergence}
\input{appdx_items/rebuttal/train_curve.tex}
In Figure ~\ref{fig:curve}, we show that training MAE on document and web screenshot images with the normalized pixel loss leads to training divergence.

\section{Stage 2: Supervised Pretraining}
\label{ssl:impl}
In this section, we present the detail implementation and the dataset composition of our supervised pretraining after the self-supervised pretraining.
\subsection{Implementation Details}
\input{appdx_items/supervised/super_hparams.tex}
We use the official LLaVA-Next~\citep{li2024llava} repo with some customization to support different LLM and encoder settings, and we use 32 H200 GPUs for the training. To handle high resolution images, we use AnyRes image tilting with size of 384x384. The hyperparameters are listed in Table~\ref{tab:super-hparams}.

\subsection{Datasets}
\input{appdx_items/supervised/super_data.tex}
We use a wide range of publicly available data for our supervised pretraining. Notably, all the data excluding UGround are reformulated into layout and information extraction. For more detail on the data curation, refer to Granite Vision~\citep{team2025granite}. For the Self-curated Data, we use PaddleOCR as our OCR model. We first filter arXiv papers to include only those published in venues or journals, ensuring a high-quality source corpus. During post-processing, we remove all bounding boxes with confidence scores below 0.5 to ensure accurate OCR localization. We then randomly resize, crop, and add noise to prevent the model from overfitting to a specific PDF size. Next, we normalize all bounding box coordinates to the range [0, 999] with respect to the new size. To control task difficulty, we discard bounding boxes that contain either too few words (e.g., single tokens) or excessively long paragraphs. Finally, from each remaining (category, bounding box, text) tuple, we construct diverse training examples for bbox-to-text, text-to-bbox, and bbox-to-category tasks. The data composition is shown in Table~\ref{tab:super-data}.

\subsection{Weight-Merging}
For the weight-merging, we randomly sample 10k document and web images from self-supervised pretraining as our training data. We use the AdamW Optimizer~\citep{loshchilov2017decoupled}, a learning rate of 1e-4, and a batch size of 32 to train for 20 epoches. The only trainable component is the newly introduced merge coefficient, while all the vision encoders remain frozen.

\section{Classic Document and web Tasks}
\label{appx:cls_doc}
\subsection{DocBank}
DocBank is a large-scale benchmark for document layout analysis, providing token-level annotations that capture both textual and structural information. It is widely used to evaluate models on tasks such as text detection, segmentation, and understanding of document formatting. In our work, we formulate DocBank as Document element recognition, where the task is the predict the bounding box of a specific element given a document image.

\input{appdx_items/cls_doc/docbank_result.tex}
\minisection{Implementation}
DocBank consists of 13 categories. For each category, we finetune the vision encoder with an attention pooling network on the corresponding training set, and perform evaluation on the validation set. Before training, each image is processed and resized according to the default preprocessor for the specific vision encoder. For each category, we train with 5 epoch, a learning rate of 1e-5, weight decay of 1e-4, and a batch size of 32. We use Smooth L1 objective~\citep{huber1964robust} for the training. The per-category result is shown in Table~\ref{tab:docbank-category}.

\subsection{Doclaynet}
DocLayNet is a large-scale dataset for document layout analysis that contains richly annotated page images collected from diverse sources such as scientific articles, reports, and business documents. It provides pixel-level segmentation masks for a broad set of layout elements, enabling fine-grained evaluation of models on tasks like detection, segmentation, and structural understanding. In our work, we use Doclaynet to perform semantic segmentation.

\minisection{Implementation}
Doclaynet consists of 11 distinct classes, and in our implementation, we introduce an additional class for the area with no class label. We finetune the vision encoder along with an MLP projector to project the visual feature patches into a segmentation mask. Then we use crossentropy~\citep{shannon1948mathematical} between the prediction and the ground truth segmentation as objective and train on the training set. We train for 2 epoch, with a learning rate of 1e-5, weight decay of 1e-4, and a batch size of 32.

\minisection{Ablation on the generalist encoder}
In this section, we conduct an ablation on the improvement to RICO-SCA when we incorporate the generalist encoder feature. As shown in Table~\ref{tab:rico-sca}, it is clear tha the generalist encoder provide high-level semnatic features that are important for classification tasks.
~\input{appdx_items/rebuttal/rico_ablation.tex}
\subsection{RICO-SCA}
RICO-SCA is a benchmark derived from the RICO dataset, which contains large-scale mobile application user interfaces. The SCA (Semantic Component Annotation) extension enriches each screen with detailed semantic labels for UI elements such as class type, buttons, images, text fields, and navigation components. In our work, we use RICO-SCA to evaluate on web UI classification.

\minisection{Implementation}
We finetune the vision encoder together with an attention pooling network to classify the class type of a given web UI image, using the provided training set. We train for 3 epoch, a learning rate of 1e-5, a weight decay of 1e-4, and a batch size of 32. We then evaluate the finetuned model on the validation set.

\section{VLM as Evaluator}
\label{appx:vlm_eval}
\input{appdx_items/vlm_eval/data.tex}
Here we provide the detail implementation and data composition on using VLMs as evaluator for vision encoders. 

\minisection{Data Composition}
We use a set of instruction tuning data with a focus on document and web understanding for the VLM training. We then use the Mind2Web training set to finetune our VLM on web agentic tasks. The full data composition can be found in Table~\ref{tab:vlm-data}

\minisection{Implementation}
The training is conducted with 32 H200 GPUs. We use the same architecture as described in Section~\ref{ssl:impl}. For image tilting, we use tilting to split the image into size of 336x336 for AIMv2, and 384x384 for other vision encoders. The instruction tuning hyperparameter and finetuning is the same as in~\ref{tab:super-hparams}, except that Mind2Web is finetuned for 2 epoch.

\subsection{Evaluation Standard Errors}
\input{appdx_items/rebuttal/stderr.tex}
In this section, we provide the standard errors for most vision-language benchmarks. As shown in Table ~\ref{tab:stderr}, the standard errors remain are reasonable, indicating that the model’s performance is stable and consistent across benchmarks. 

\subsection{Evaluation on Cross-Domain Tasks}
\input{appdx_items/rebuttal/cross_domain.tex}
Here, we provide results on DTCBench for Korean document understanding and CMMMU for Chinese visual question answering. In Table ~\ref{tab:cmmmu-dtc}, DAVE's consistent improvements on these benchmarks show that it can generalist to cross domain

\section{Ablations}
\label{appx:add_abl}
In this section, we provide the implementation and evaluation details of our ablation studies, along with additional ablations on \NAME.
\subsection{Inter-Patch Variance}
We randomly sample 1000 document and web images from our pretraining data to conduct the analysis. 
For each image, we divide it into non-overlapping $16 \times 16$ patches and compute the per-patch standard deviation of pixel intensities after normalization. 
This provides a distribution of inter-patch variance that reflects the inherent local variability in different domains. 
As discussed in Section~\ref{abl}, document-like images exhibit consistently lower variance across patches, while web-like and natural images demonstrate a broader spread. 
These differences highlight the difference in low-level representation, particularly structural and spatial, between natural and document/web images.

\subsection{Multi-Encoder setup}
Given a setup consisting of SigLIP~2 and a specialized encoder, the images are first processed separately by their respective preprocessor. The processed images are then passed into the corresponding vision encoders to obtain visual patch features. Finally, the patch features are concatenated channel-wise. In scenarios where the specialized vision encoder produces a different number of patch features compared to SigLIP~2, we apply linear interpolation on the specialized features to match the patch count.

\input{appdx_items/ablation/abl_aimv2.tex}

\subsection{Generalization to Unseen Generalist Encoder}
We denote the specialized encoder within \NAME as \NAME-spec.
To evaluate whether \NAME-spec can adopt to unseen generalist encoder, we pair it with with AIMv2 during instruction
tuning. The result is show in Table~\ref{tab:abl-aimv2}. The comparable performance with DAVE highlights the generalizability of DAVE-spec to novel vision-language and agent frameworks.

%% file: appdx_items/ssl/ssl_hparams.tex
\begin{table}[t]
\centering
\footnotesize
\begin{tabularx}{0.48\textwidth}{@{} l N @{}}
\toprule
\textbf{Hyperparameter} & \textbf{Value} \\
\midrule
Batch Size     & 4096 \\
Learning Rate   & 1e-5 \\
Epochs          & 25 \\
Warmup Epochs   & 5 \\
Weight Decay    & 0.05 \\
Mask Ratio      & 0.75 \\
LR Scheduler    & Cosine \\
\bottomrule
\end{tabularx}
\caption{Training hyperparameters for MAE training.}
\label{tab:ssl-hparams}
\end{table}

%% file: appdx_items/ssl/ssl_result.tex
\begin{table}[h]
\centering
\footnotesize
\begin{tabularx}{0.48\textwidth}{@{} l *{3}{N} @{}}
\toprule
\textbf{Model} & \textbf{DocBank} & \textbf{Doclaynet} & \textbf{RICO-SCA} \\
\midrule

MAE-Web &  64.6& 44.5& 88.3\\
MAE-Doc &  72.8&52.5 & 90.7\\
\bottomrule
\end{tabularx}
\caption{Evaluation results on classic document tasks.}
\label{tab:ssl-result}
\end{table}

%% file: appdx_items/rebuttal/train_curve.tex
\begin{figure}[t]
    \centering
    \includegraphics[width=0.5\linewidth]{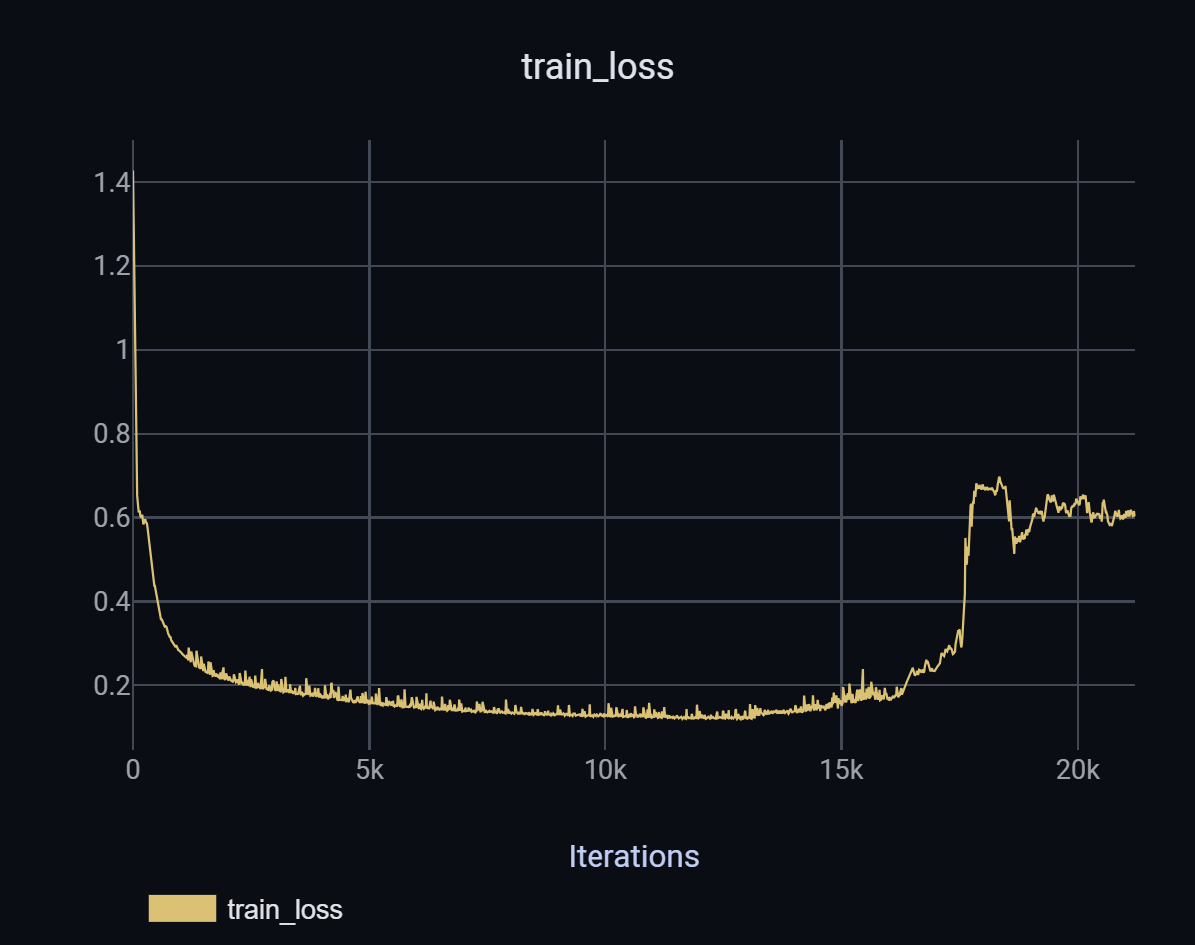}
    \caption{
    Training loss curve of MAE with normalized pixel as objective.
    }
    \label{fig:curve}
\end{figure}

%% file: appdx_items/supervised/super_hparams.tex
\begin{table}[t]
\centering
\footnotesize
\begin{tabularx}{0.48\textwidth}{@{} l N @{}}
\toprule
\textbf{Hyperparameter} & \textbf{Value} \\
\midrule
Max sequence length   & 20000 \\
Learning Rate   & 3e-5 \\
Epochs          & 1 \\
Warmup ratio   & 0.03 \\
LR Scheduler    & Cosine \\
\bottomrule
\end{tabularx}
\caption{Training hyperparameters for supervised pretraining.}
\label{tab:super-hparams}
\end{table}

%% file: appdx_items/supervised/super_data.tex
\begin{table}[h]
\centering
\footnotesize
\begin{tabularx}{0.9\textwidth}{@{} l N X @{}}
\toprule
\textbf{Dataset} & \textbf{\#Image} & \textbf{Description} \\
\midrule
FM4D & 250k & Doc/Chart/Table Extraction \\
PlotQA & 146k & Plot Extraction \\
Fintabnet & 88k & Table Extraction \\
ChartQA & 17k & Chart Extraction \\
Datikz & 94k & Image to Latex \\
Pubtables & 480k & Table to HTML \\
UGround & 756k & Web UI Grounding \\
Self-curated Data & 250k & Doc Grounding/structure Recognition \\
\bottomrule
\end{tabularx}
\caption{Dataset description for supervised pretraining.}
\label{tab:super-data}
\end{table}

%% file: appdx_items/cls_doc/docbank_result.tex
\begin{table*}[t]
\centering
\footnotesize
\begin{tabularx}{\textwidth}{@{} l *{6}{N} @{}}
\toprule
\textbf{Model} & \textbf{Abstract} & \textbf{Author} & \textbf{Caption} & \textbf{Equation} & \textbf{Figure} & \textbf{Footer} \\
\midrule

DinoV2     & 86.4 & 9.0  & 45.2 & 14.9 & 90.5 & 3.6  \\
MAE        & 84.1 & 17.9 & 20.7 & 33.4 & 90.8 & 23.8 \\
Dolphin    & 73.8 & 21.6 & 48.1 & 25.7 & 91.5 & 32.9 \\
Pix2Struct &   80.6   &   26.3   &  21.6    &  15.3    &83.1      & 27.7     \\
AIMv2      & 85.8 & 5.5  & 27.7 & 15.5 & 90.4 & 19.4 \\
SigLIP 2   & 85.9 & 23.1 & 52.6 & 34.7 & 91.6 & 31.6 \\
\NAME      &   87.4   &  35.3    &  54.1    &   53   &  91.5    & 17.9     \\
\midrule
\end{tabularx}

\begin{tabularx}{\textwidth}{@{} l *{6}{N} @{}}
\textbf{Model} & \textbf{List} & \textbf{Paragraph} & \textbf{Reference} & \textbf{Section} & \textbf{Table} & \textbf{Title} \\
\midrule
DinoV2     & 42.2 & 10.3 & 86.1 & 3.2 & 68.5 & 0.0  \\
MAE        & 40.8 & 10.2 & 88.8 & 9.2 & 88.1 & 26.5 \\
Dolphin    & 68.7 & 11.5 & 88.0 & 9.7 & 84.3 & 50.2 \\
Pix2Struct &   33.9   &   13.7   &  87.3    &  5.4   &  62.2    & 64.4     \\
AIMv2      & 45.8 & 9.8  & 86.2 & 2.8 & 80.3 & 39.0 \\
SigLIP 2   & 51.8 & 13.2 & 88.7 & 7.5 & 90.9 & 48.8 \\
\NAME      &   66.4   &  20.6    &   91.6   &  27   &   87.9   & 49.7     \\
\bottomrule
\end{tabularx}

\caption{Evaluation results on DocBank by category.}
\label{tab:docbank-category}
\end{table*}

%% file: appdx_items/rebuttal/rico_ablation.tex
\begin{table}[h]
\centering
\footnotesize
\begin{tabularx}{0.48\textwidth}{@{} l N @{}}
\toprule
\textbf{Model} & \textbf{RICO-SCA} \\
\midrule
SigLIP 2         & 93.3 \\
DAVE-specialized & 90.7 \\
DAVE             & 92.3 \\
\bottomrule
\end{tabularx}
\caption{Performance on the RICO-SCA benchmark.}
\label{tab:rico-sca}
\end{table}

%% file: appdx_items/vlm_eval/data.tex
\begin{table}[h]
\centering
\footnotesize
\begin{tabularx}{0.8\textwidth}{@{} l N X @{}}
\toprule
\textbf{Dataset} & \textbf{\#Image} & \textbf{Description} \\
\midrule
AI2D & 2k &  Chart understanding\\
DocVQA & 10k & Document understanding \\
InfoVQA & 2k & Infographic reasoning \\
ChartQA & 17k &  Chart understanding\\
Pixmo-Doc & 250k & Document understanding \\
MultiUI & 2m &  Web localization and understanding\\
LLaVA 665K & 340k & General visual instruction data \\
\bottomrule
\end{tabularx}
\caption{Dataset description for VLM training.}
\label{tab:vlm-data}
\end{table}

%% file: appdx_items/rebuttal/stderr.tex
\begin{table}[h]
\centering
\footnotesize
\begin{tabularx}{0.9\textwidth}{@{} l *{6}{N} @{}}
\toprule
\textbf{Model} & \textbf{AI2D} & \textbf{ChartQA} & \textbf{DocVQA} & \textbf{InfoVQA} & \textbf{RealWorldQA} & \textbf{TextVQA} \\
\midrule
llama-Siglip & 0.88 & 0.99 & 0.55 & 0.85 & 1.80 & 0.64 \\
llama-DAVE   & 0.88 & 0.97 & 0.47 & 0.87 & 1.77 & 0.62 \\
qwen-Siglip  & 0.79 & 0.85 & 0.36 & 0.74 & 1.70 & 0.58 \\
qwen-DAVE    & 0.78 & 0.76 & 0.35 & 0.76 & 1.80 & 0.59 \\
\bottomrule
\end{tabularx}
\caption{Standard errors in vision-language benchmarks.}
\label{tab:stderr}
\end{table}

%% file: appdx_items/rebuttal/cross_domain.tex
\begin{table}[h]
\centering
\footnotesize
\begin{tabularx}{0.48\textwidth}{@{} l *{2}{N} @{}}
\toprule
\textbf{Model} & \textbf{CMMMU} & \textbf{DTCBench} \\
\midrule
SigLIP 2 & 21.4 & 29.2 \\
DAVE     & 26.1 & 38.3 \\
\bottomrule
\end{tabularx}
\caption{Evaluation results on CMMMU and DTCBench.}
\label{tab:cmmmu-dtc}
\end{table}

%% file: appdx_items/ablation/abl_aimv2.tex
\begin{table}[h]
\centering
\footnotesize
\begin{tabularx}{0.48\textwidth}{@{} l *{2}{N} @{}}
\toprule
\textbf{Model} & \textbf{Doc} & \textbf{Web} \\
\midrule

\NAME &  63.4& 68.2\\
\NAME-spec + AIMv2 &  64.5&64.8\\
\bottomrule
\end{tabularx}
\caption{Evaluation results on document and web understanding.}
\label{tab:abl-aimv2}
\end{table}